\documentclass{article}
\usepackage{spconf,amsmath,graphicx}

\usepackage{amssymb}
\usepackage{amsfonts}
\usepackage{bm}
\usepackage{todonotes}
\usepackage{algorithm}
\usepackage{algpseudocode}
\usepackage{subcaption}
\usepackage[capitalize]{cleveref}
\usepackage{color}
\usepackage{cleveref}
\usepackage{autonum}

\title{Attention Mechanism in Randomized Time Warping}

\name{Yutaro Hiraoka$^{\star}$ \qquad Kazuya Okamura$^{\dagger}$ \qquad Kota Suto$^{\dagger}$ \qquad Kazuhiro Fukui$^{\ddagger}$}
\address{$^{\star}$ The Japan Research Institute, Limited \\
  $^{\dagger}$Graduate School of Science and Technology, University of Tsukuba \\
  $^{\ddagger}$Tsukuba Institute for Advanced Research, Department of Computer Science, University of Tsukuba}

\begin{document}
\maketitle
\begin{abstract}
This paper reveals that we can interpret the fundamental function of Randomized Time Warping (RTW) as a type of self-attention mechanism, a core technology of Transformers in motion recognition. 
The self-attention is a mechanism that enables models to identify and weigh the importance of different parts of an input sequential pattern. On the other hand, RTW is a general extension of Dynamic Time Warping (DTW), a technique commonly used for matching and comparing sequential patterns.
In essence, RTW searches for optimal contribution weights for each element of the input sequential patterns to produce discriminative features. Although the two approaches look different, these contribution weights can be interpreted as self-attention weights. In fact, the two weight patterns look similar, producing a high average correlation of 0.80 across the ten smallest canonical angles. 
However, they work in different ways: RTW attention operates on an entire input sequential pattern, while self-attention focuses on only a local view which is a subset of the input sequential pattern because of the computational costs of the self-attention matrix. This targeting difference leads to an advantage of RTW against Transformer, as demonstrated by the 5\% performance improvement on the Something-Something V2 dataset.
\end{abstract}

\renewcommand{\thefootnote}{\fnsymbol{footnote}}
\footnote[0]{
\copyright 2025 IEEE. Personal use of this material is permitted. Permission from IEEE must be obtained for all other uses, in any current or future media, including reprinting/republishing this material for advertising or promotional purposes, creating new collective works, for resale or redistribution to servers or lists, or reuse of any copyrighted component of this work in other works.
}
\renewcommand{\thefootnote}{\arabic{footnote}}

\section{Introduction}
\label{sec:intro}
In this paper, we reveal that the essential mechanisms of Randomized Time Warping (RTW) \cite{rtw} can be regarded as a type of self-attention weights used as a core component of Transformer \cite{transformer}. The self-attention weights effectively generate discriminative features from an input sequential pattern in motion recognition. 
In contrast, RTW is a general extension of dynamic time warping (DTW), commonly used as the standard technique for matching two given sequential patterns. RTW is robust against changes in motion speed and has a high discrimination ability even when training data is scarce \cite{eGDS,TSTF}. It has been used effectively for motion recognition, such as gesture and action \cite{eGDS,TSTF}. 

In the following, we provide an overview of the mechanisms of self-attention in Transformer and RTW, respectively.
Self-attention is well-known as a fundamental mechanism that allows models to identify and weigh the importance of different parts of an input sequential pattern in Transformers, such as in natural language processing (NLP) \cite{transformer} and computer vision tasks \cite{vit, timesformer}. We can obtain such importance for each element as a self-attention weight matrix. This matrix consists of the similarities between the Query and Key vectors calculated by applying different linear transformations to the input sequential pattern. Using this self-attention weight matrix, we obtain a discriminative feature vector as the weighted sum of the Value vectors, which are also generated using another linear transformation to the input sequential pattern.   

\begin{figure}[tb]
  \centering
     \includegraphics[width=87mm]{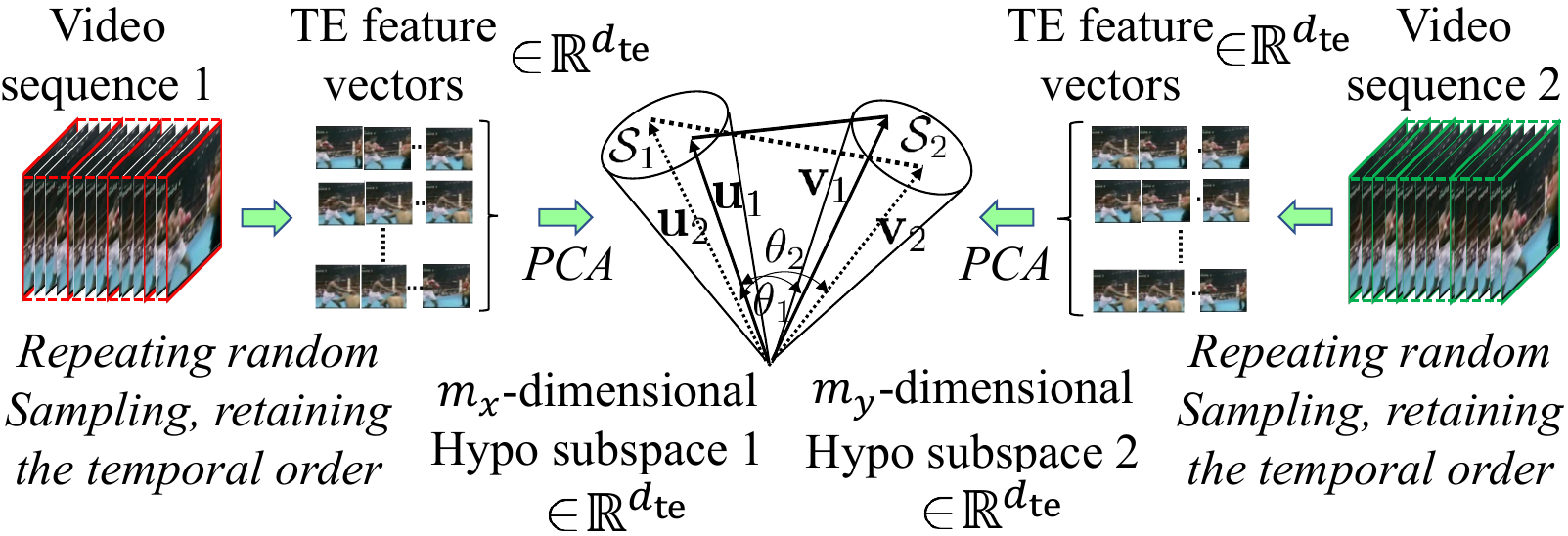}
     \caption{Framework of Randomized Time Warping (RTW). Two sets of input and reference TE features are represented by a hypo subspace, respectively. RTW measures the structural similarity between the $k$-dimensional hypo subspaces, ${\cal{S}}_1$ and ${\cal{S}}_2$, by using the $k$ canonical angles ${\{\theta_i\}}_{i=1}^k$ between them, as the structural correlation between two sets of TE features. Each canonical angle $\theta_i$ is formed by a pair of canonical vectors $\mathbf{u}_i$ and $\mathbf{v}_i$.}
     \label{fig:RTWsubspace}  
     \end{figure}

On the other hand, the essential mechanism of RTW is to simultaneously calculate the similarities between many pairs of warped patterns referred to as Time Elastic (TE) features. TE features are generated by randomly sampling the input sequential pattern while retaining their temporal order from the two sequential patterns as shown in \cref{fig:RTWsubspace}.
Instead of using DTW to search for the most similar warped patterns from many sequential patterns, RTW compactly represents the set of TE features as a low-dimensional subspace called the hypo subspace. Then, RTW measures the structural similarity using multiple canonical angles between two hypo subspaces generated from two sequences. 
This subspace representation converts the high-cost computational process of calculating the similarity between many pairs of TE features to a light and straightforward calculation of structural similarity between two hypo subspaces. The above two-step process is essential for finding optimal contribution weights for each element of the input sequential patterns to produce discriminative features like the self-attention module. We call the contribution weight the ``RTW attention pattern''.

Although these two approaches might look different, the RTW attention pattern corresponds well to the self-attention pattern of the Transformer. As shown in the experiments section, the average correlation between the two attention patterns reaches a high value of 0.80 on the large-scale motion recognition dataset, Something-Something V2. 

However, in practical recognition tasks, RTW attention and self-attention modules operate in different ways. RTW attention is applied uniformly across all input sequential patterns, while self-attention modules focus solely on a specific local view (short video clip) of these input sequential patterns. The size of the self-attention matrix increases with the squared length of a targeted video clip, resulting in higher computational costs.
This difference leads to an advantage of RTW over Transformer, demonstrating a performance improvement of 5\% on the Something-Something V2 dataset.
We also found that RTW only uses 10\% of the entire dataset in order to achieve the same performance level as the Transformer trained on the complete dataset. Our findings suggest that RTW could be a viable alternative to the self-attention module.

The main contributions of this paper are as follows:
\begin{itemize}
  \item[(1)] We reveal that the essential mechanism of RTW, searching for optimal contribution weights of each element of an input sequential pattern, can be regarded as a type of self-attention. We demonstrate the similarity between RTW attention and Transformer self-attention.
    \item[(2)] We discuss that the RTW attention works on an entire sequential pattern. In contrast, self-attention works on only a local view which is a subset of the sequential pattern. As a result, we demonstrated that the RTW outperforms Transformers with self-attention on the Something-Something V2 dataset.  
\end{itemize}

\section{Randomized Time Warping}
\label{sec:rtw}
In this section, we provide an overview of the mechanism of RTW according to \cref{fig:RTWmsm}, the critical components of RTW, Time Elastic (TE) features and hypo subspace. 

\subsection{Time Elastic features}
Given input sequential patterns that consist of ${N}$ features, ${\{{\mathbf{z}}_i \}}_{i=1}^{N}$, which a DNN model extracts from an input video.
The critical component of RTW is the concept of a time-warped pattern called Time Elastic (TE) feature, which is generated by randomly selecting ${R}$ features from the input sequential patterns, while retaining its temporal order, as shown in \cref{fig:RTWmsm}. 
A large number of TE features ${\{ {\mathbf{f}}_i \}}_{i=1}^L \in \mathbb{R}^{d_\text{te}}$, where $d_\text{te}={d_\text{model}}{\times}{R}$, are collected by $L$ sampling from the input sequential patterns. They contain many kinds of time-warped patterns from local to global scales. Similarly, a set of various TE features ${\{ {\mathbf{g}}_i \}}_{i=1}^L \in \mathbb{R}^{d_\text{te}}$ is generated from each reference sequence. 

\subsection{Generation of hypo subspace}
Computing the correlation among all possible linear combinations of TE features ${\{\mathbf{f}_i\}}_{i=1}^L$ is not practical. Therefore, RTW compactly represents a set of TE features by a low-dimensional subspace called the hypo subspace, as shown in \cref{fig:RTWmsm}.
Hypo subspaces $\mathcal{S}_1$ and $\mathcal{S}_2$ are generated by applying the principal component analysis (PCA) to sets of TE features ${\{\mathbf{f}_i\}}_{i=1}^L$ and ${\{\mathbf{g}_i\}}_{i=1}^L$, respectively. 
Given ${\mathbf{F}}=[\mathbf{f}_1,\cdots, \mathbf{f}_L] \in \mathbb{R}^{d_\text{te} \times L}$ and ${\mathbf{G}}=[\mathbf{g}_1,\cdots, \mathbf{g}_L]  \in \mathbb{R}^{d_\text{te} \times L}$, the orthonormal basis matrices $\mathbf{X}$ and $\mathbf{Y}$ of hypo subspaces $\mathcal{S}_1$ and $\mathcal{S}_2$ can be obtained as the eigenvectors corresponding to the $m$ largest eigenvalues of the matrices ${\mathbf{F}}{\mathbf{F}}^T$
and ${\mathbf{G}}{\mathbf{G}}^T$, respectively.

\subsection{Structural similarity between two hypo subspaces}
The structural similarity $sim(\mathcal{S}_1,\mathcal{S}_2)$ between $m_x$-dimensional $\mathcal{S}_1$ and $m_y$-dimensional $\mathcal{S}_2$ can be measured with multiple canonical angles ${\{\theta_i\}}_{i=1}^{m_x}$ between them under the mutual subspace method (MSM) framework \cite{msm2, gds} and $m_x \leq m_y$. The directions of the canonical angles are orthogonal to each other.  

Let $\mathbf{X} \in \mathbb{R}^{d_\text{te} \times m_x}$ and $\mathbf{Y} \in \mathbb{R}^{d_\text{te} \times m_y}$ be the orthonormal basis vectors of the two hypo subspaces $\mathcal{S}_1$ and $\mathcal{S}_2$, with $m_x$ and $m_y$ dimensions, respectively.
We calculate Singular Value Decomposition (SVD) ${\mathbf{X}}^{\top}{\mathbf{Y}} = {\mathbf{U}} {\bm{\Sigma}} {\mathbf{V}}^{\top}$, where $diag(\bm{\Sigma}) = (\kappa_{1},\ldots,\kappa_{m_p})$,
$\{\kappa_{i} \}_{i=1}^{m_p}$ represents the set of singular values($={\rm{cos}}{\theta_i}={\mathbf{u}}_i^t{\mathbf{v}}_i$), and $\kappa_{1} \geq \ldots \geq \kappa_{m_p}$. 
The similarity can then be defined  as $ Sim({\mathcal{S}}_1, {\mathcal{S}}_2) = \frac{1}{r}\sum_{i=1}^{r} \kappa_i^2$, where $1 \leq r \leq m_x$.

\subsection{Classification algorithm of Randomized Time Warping}
RTW is a two-stage procedure, divided into learning and inference stages. 
Given a set of discriminate feature vectors, such as deep features, extracted from sequential images, we consider classifying an input video into $C$ class. Each class has $n_c$ videos.

\vspace{1mm}
{\bf{Learning stage:}}
(1) Randomly selected ${R}$ images, while retaining their temporal order, from a given video are fed into a DNN model like a spatial feature extraction module. This results in the extraction of a subset of feature vectors, generating the TE feature vector by concatenating those feature vectors.
(2) The process in (1) is repeated $L$ times for the video of the $c$-th class to obtain a set of TE feature vectors ${\{{\mathbf{f}}_i^c\}}_{i=1}^L$.
(3) $N_{c}$-dimensional reference hypo subspace ${\mathcal{S}}_c^i$ is generated by applying PCA to the set of TE feature vectors ${\{{\mathbf{f}}_i^c\}}_{i=1}^L$. In this way, we obtain a set of all reference hypo subspaces ${\{{\mathcal{S}}_c^i\}}_{i=1}^{n_c}$ for all $C$ classes.

\vspace{1mm} 
{\bf{Inference stage:}}
(1) A set of TE feature vectors ${\{\mathbf{g}_i\}}_{i=1}^{L}$ is generated from an input video in the same procedure used in the learning process. 
(2)  $N_{in}$-dimensional input hypo subspace $\mathcal{S}_{\text{in}}$ is generated by applying PCA to the set of TE feature vectors, ${\{\mathbf{g}_i\}}_{i=1}^{L}$.   
(3)  For the $c$-th class, the similarities between $\mathcal{S}_{\text{in}}$ and 
    ${\{{\mathcal{S}}_c^i\}}_{i=1}^{n_c}$ are measured by using the MSM. 
(4)  The input video is classified into the correct class using $k$-nearest neighbors, where we define the similarity for the $c$-th class as the average of the top $k$ similarities of the input hypo subspace and multiple reference hypo subspaces of that class.

\begin{figure}[bt]
  \centering
    \includegraphics[width=84mm]{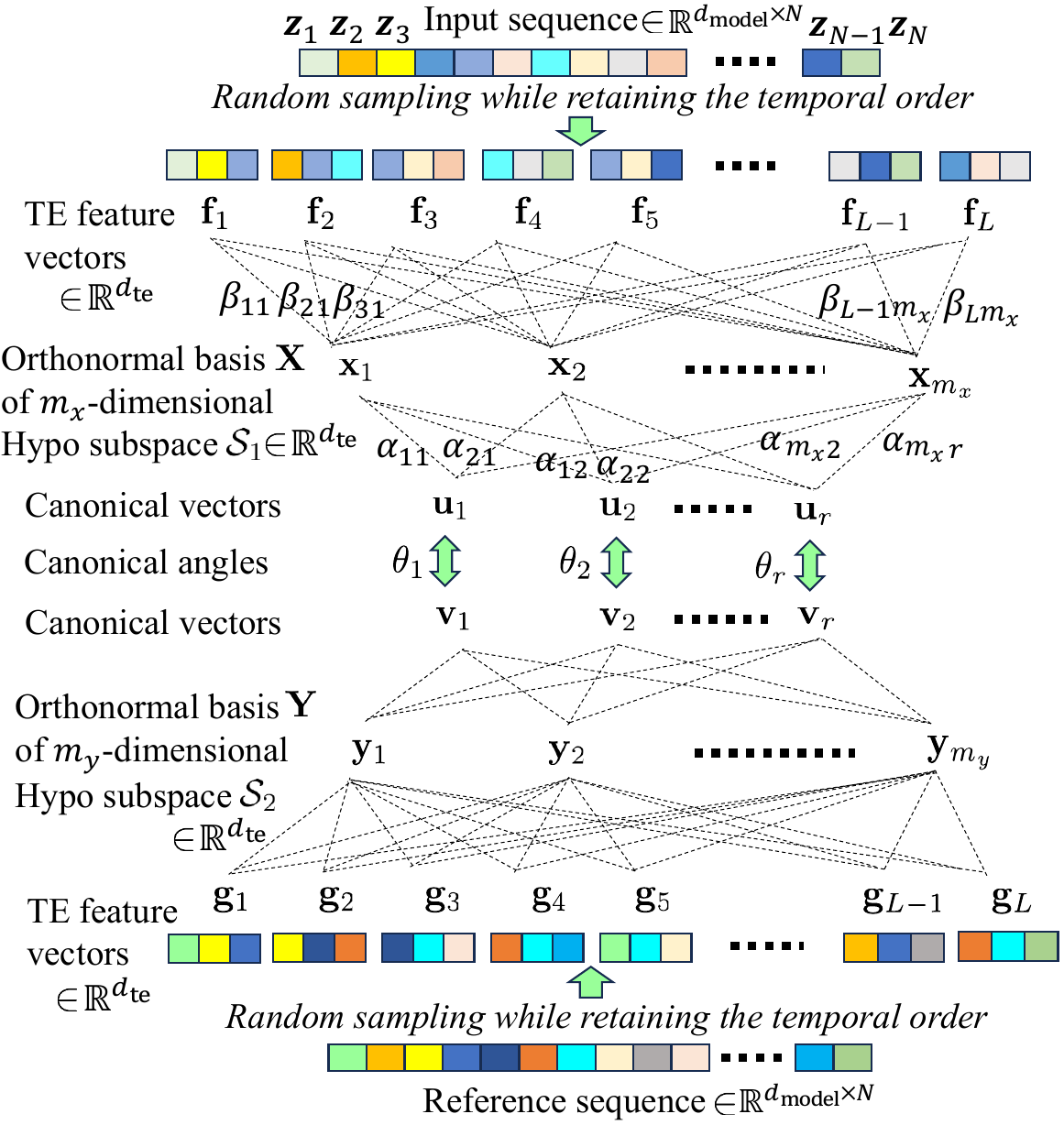}  
  \caption{Calculation of contribution weights for linear combination of TE features in the process of RTW.}
  \label{fig:RTWmsm}  
\end{figure}
\section{Attention Patterns of RTW and self-attention module}
\label{sec:correspondence}
In this section, we explain the computation process for each attention pattern of RTW and self-attention module in Transformer in detail. 

\subsection{Computation of Self-attention Pattern in Transformer}
The main process of self-attention is summarized as follows:
First, self-attention creates three components: query vectors $\{\mathbf{q}_i\}_{i=1}^N$$ \in  R^{d_k}$, key vectors $\{\mathbf{k}_i\}_{i=1}^N$$ \in  R^{d_k}$, and value vectors $\{\mathbf{v}_i\}_{i=1}^N$$ \in  R^{d_v}$ from the $i$-th input component vector of an input pattern consisting of $N$ sequential vectors $\{\mathbf{I}_i\}$ $ \in  R^{d_\text{model}}$.
These vectors, $\{\mathbf{q}_i\}$, $\{\mathbf{k}_i\}$, and $\{\mathbf{v}_i\}$, are obtained by applying the input sequential vectors to learnable transformation matrices $\mathbf{W}_Q$, $\mathbf{W}_K$, and $\mathbf{W}_V$, respectively.
Next, the self-attention weights $\{{\mathbf{A}_{ij}}\}_{j=1}^{N}$ for the input vector $\mathbf{I}_i$ is calculated as the dot product of the vector $\mathbf{q}_i$ with the vectors $\{{\mathbf{k}_j}\}_{j=1}^{N}$, divided by the square root of the dimension $d_k$ and a Softmax function is applied. A self-attention weight matrix $\mathbf{A}$ is obtained by applying this operation to each of the query vectors.
The self-attention weight ${\mathbf{A}_{ij}}$ represents the attention from the $i$-th element of the input sequence to the $j$-th element, indicating how much it contributes to the generation of the discriminative feature vector $\mathbf{r}_i$$ \in  R^{d_\text{model}}$ in motion classification.
Finally, the output discriminative feature vectors $\{\mathbf{r}_i\}$ are computed as the weighted sum of the value vectors; the $i$-th weighted sum of the value vectors is obtained by using $\{\mathbf{v}_j\}$$ \in  R^{d_v}$ with the self-attention weights $\{{\mathbf{A}_{ij}}\}_{j=1}^{N}$.

We obtain the sum of attention weights directed to each element of the input sequence as $s_j = \sum_{i=1}^{N}{\mathbf{A}_{ij}}$.
We define the attention weight pattern as the vector ${\mathbf{s}=}$[$s_1$, \dots ,$s_N$].

\begin{figure}[t]
  \begin{center}
    \includegraphics[width=87mm]{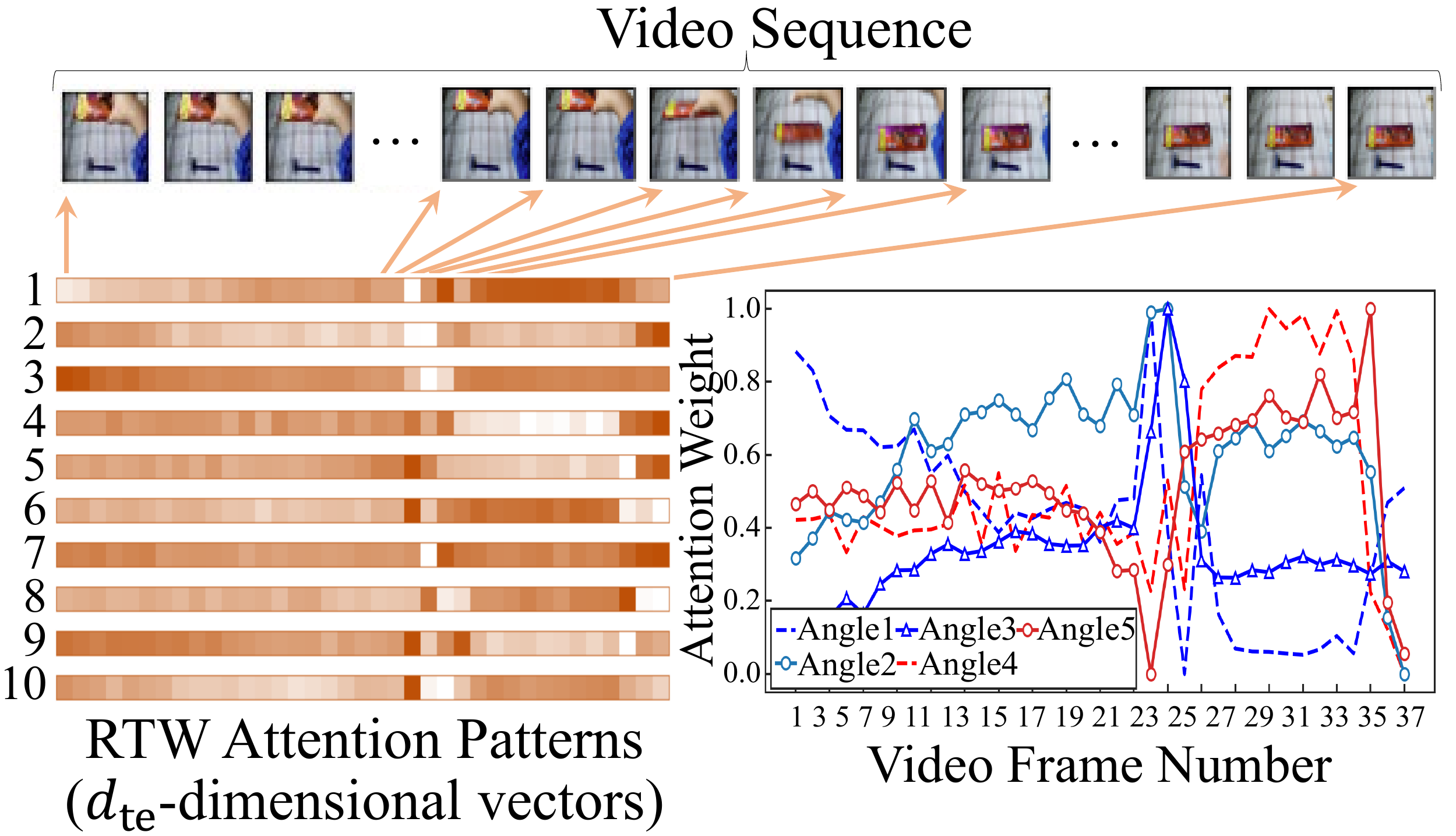}
  \end{center}
  \caption{An example of RTW attention patterns corresponding to ten canonical angles. This is for a video of ``dropping a biscuit behind a razor.'' In this heat map, lighter colors indicate greater contribution. The line chart on the right shows the top five contribution weights in the same model.}
  \label{fig:rtw_att_sample}  
\end{figure}

\subsection{Computation of RTW attention patterns}
We explain the procedure for calculating the attention patterns of an input sequence in RTW, according to \cref{fig:RTWmsm}.  Given a set of $N$ sequential deep feature vectors $\{{\mathbf{z}}_i\}_{i=1}^N$ and then a set of $L$ TE features $\{\mathbf{f}_i\}_{i=1}^L$ were generated from $\{{\mathbf{z}}_i\}_{i=1}^N$, we consider the attention pattern regarding the $k$-th canonical vector ${\mathbf{u}}_{k}$ forming the $k$-th canonical angle $\theta_k$. \cref{fig:rtw_att_sample} shows an example of RTW attention patterns.

\begin{itemize} 
  \item[1)] [{\bf TE features $\rightarrow$ Canonical vector}] We can calculate the linear combination weight ${{w}}_{ik}$ of the $i$-th TE feature vector to the $k$-th canonical vector ${\mathbf{u}}_k$ from the linear combination coefficients, $\boldsymbol \alpha$ and $\boldsymbol \beta$, obtained in the two steps of generating hypo subspaces and conducting MSM.
Concretely, the linear combination weight ${{w}}_{ik}$ is calculated as $\sum_{m=1}^{m_x} \alpha_{mk}\beta_{im}$.
  \item[2)] {\bf[Input sequence $\rightarrow$ TE features]} We define the contribution weight ${\hat{w}}_{ji}$ of the $j$-th element $\mathbf{z}_j$ of an input sequence with the length of $N$ to the $i$-th TE feature vector as 
  the value of dividing the appearance frequency (zero or one) of the element $\mathbf{z}_j$ in the $i$-th TE features by the total appearance frequency $\mathbf{c}_j$ of the element $\mathbf{z}_j$ across all TE features.
  \item[3)] {\bf{[Input sequence $\rightarrow$ TE features $\rightarrow$ Canonical vector]}} We obtain the contribution weight $t_j$ of the element $\mathbf{z}_j$ to the $k$-th canonical vector ${\mathbf{u}}_k$ as $t_j = \sum_{i=1}^L {{\hat{w}}}_{ji}{{w}}_{ik}$.
 \item[4)] We obtain an $N$-dimensional vector, ${\mathbf{t}}^{*}=$[$t_1$, \dots ,$t_N$], and finally define its normalized vector as the attention pattern vector ${\mathbf{T}}_k=\frac{{\mathbf{t}}^{*}}{||{\mathbf{t}}^{*}||}
 $ for the canonical vector ${\mathbf{u}}_k$.
 \end{itemize}

\section{Experiments}
\label{sec:experiments}
In this section, we conduct two experiments as follows:
1) We show the close connection between RTW and the self-attention module by calculating the correlation between their attention patterns.
2) We compare the performance of RTW and Transformer to demonstrate the advantage of RTW attention over the self-attention module in video classification, using the Something-Something V2 (SSv2) \cite{ssv2} dataset.

\subsection{Evaluation dataset}
The SSv2 is a public video dataset that contains various actions and objects. Its specifications are as follows: the average frame length of sequences is 40, with a maximum length of 119 and a standard deviation of 10. The dataset comprises 193k training videos and 27k test videos across 174 classes.

\subsection{Parameter settings of Transformer and RTW}
We utilized ViViT \cite{vivit} as a framework incorporating Transformers with self-attention modules. We structured a standard two-stage architecture: Transformer \#1 for spatial feature extraction and Transformer \#2 for temporal feature extraction. This structure balances computational efficiency and accuracy, ensuring effective spatial and temporal analysis \cite{vivit,vtn}. 
In our experiments, we adopted this architecture with a standard configuration, where each Transformer block processes inputs in the order of Layer Normalization, Multi-head Self-Attention, Layer Normalization, and a Multi-Layer Perceptron (MLP), incorporating a residual connection by adding the input before Layer Normalization.

For Transformer \#1, we utilized ViT \cite{vit}, which was pre-trained on ImageNet-21k \cite{imgnet}. This model transforms RGB images of size 224$\times$224 pixels into 16$\times$16 patches, outputting 768-dimensional feature vectors. Transformer \#1 converts video frames to spatial feature vectors and feeds them to  Transformer \#2, outputting temporal feature extraction. Furthermore, the output from the second Transformer is input to the MLP, forming a two-stage architecture for motion recognition. Transformer \#2 was composed of multi-head self-attention with 10 heads. 

For the parameter settings of RTW, the number of canonical angles used was set to 10. RTW also used spatial feature vectors converted from video frames using Transformer \#1, which has the same trained weights. Other parameter settings specific to the experiments are explained in each subsection.

\subsection{Comparison between RTW attention and self-attention patterns of Transformers}
\label{sec:sim_attn}
We examine the close link between the functions of RTW and the self-attention module. To investigate this relationship, we compared RTW attention and self-attention of the Transformer by measuring the correlation (cosine similarity) between the two attention pattern vectors $\mathbf{T} \in \mathbb{R}^N$ and $\mathbf{S}\in \mathbb{R}^N$ in a straightforward experimental setup. Our analysis focused on a specific view (video clip of 16 frames) extracted from the central term of the sequential patterns for both methods. The number of random sampling frames $N$ was 8 for RTW. We generated a set of 100 TE features $\{\mathbf{f}_i\}_{i=1}^{100} \in \mathbf{R}^{16}$ by repeatedly performing random sampling. The dimension $m_x$ and $m_y$ of hypo subspace were set to 50. The Transformer dealt with the view without any modification.

The order of canonical angles in RTW shows their contribution degrees, where the first canonical angle contributes mostly to the classification process. In contrast, the order of heads in multi-head self-attention is not necessarily related to their contribution degrees. 

Thus, we needed to reorder the multiple heads based on the canonical angles. First, we selected a head with the most similar self-attention pattern to the RTW attention pattern of the first canonical angle. 
Then, we selected the next self-attention pattern, which is most similar to the RTW attention pattern corresponding to the second canonical angle, from the remaining heads. In this way, we paired all the RTW attention patterns with their most similar self-attention patterns.
Finally, we calculated the correlation for the 10 pairs obtained of attention patterns $\mathbf{T} \in \mathbb{R}^{10}$ and $\mathbf{S}\in \mathbb{R}^{10}$.

\begin{figure}[tb]
  \begin{center}
    \includegraphics[width=78mm]{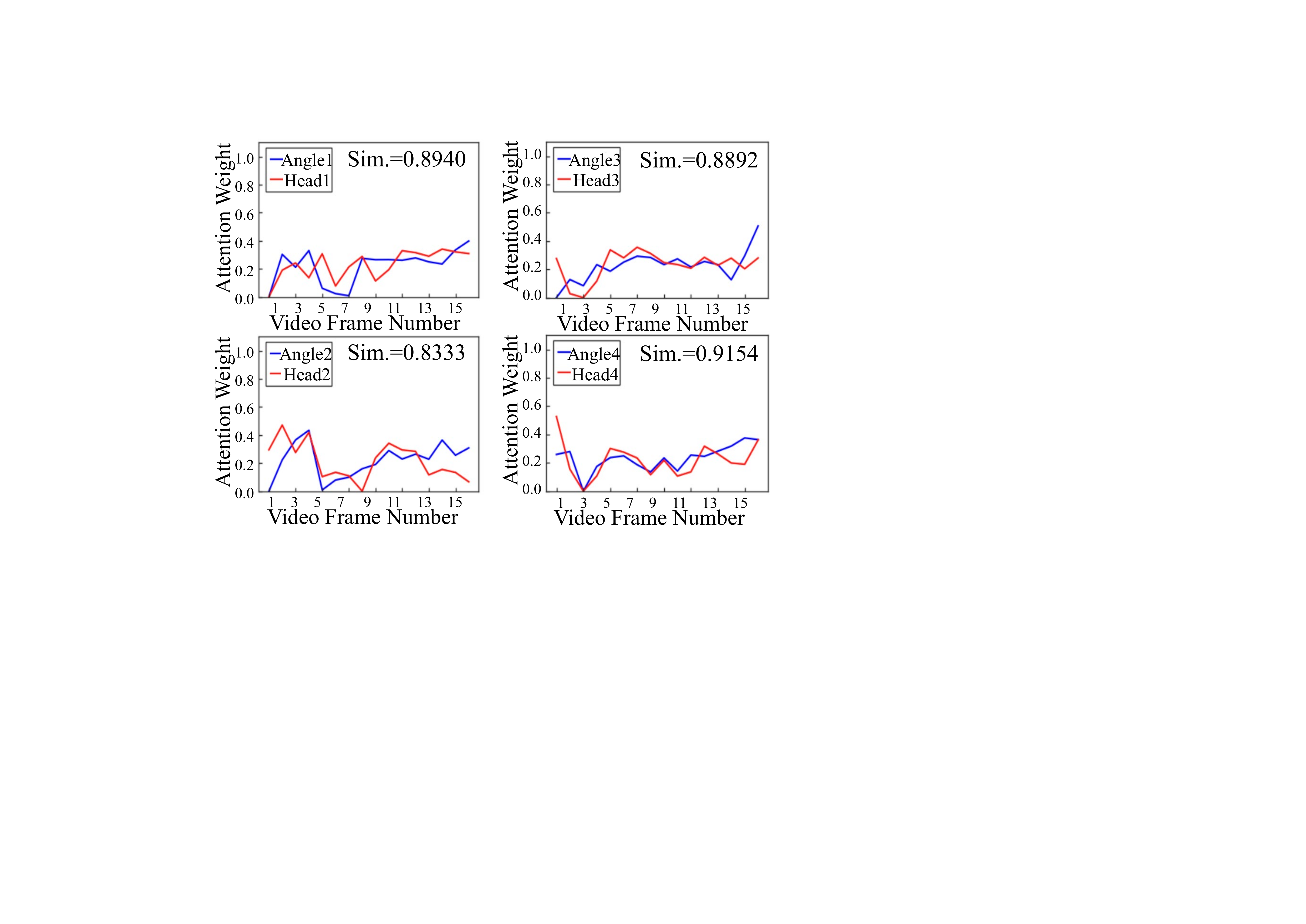}   
  \end{center}
  \caption{An example of comparison between RTW attention and self-attention patterns in terms of top four matched pairs of canonical angle and head. In the figures, "Sim." indicates the correlation (Cosine similarity) between two attentions.}
  \label{fig:pairs_cossim}  
\end{figure}

\cref{fig:pairs_cossim} shows an example of the similarity between RTW attention and self-attention patterns for the top four matched pairs of canonical angle and head.
\cref{tab:cos_sim} shows the average correlation of each pair across 174 classes. The symbol ${\mathbf{u}}_k$ in the table indicates the $k$-th canonical angle, and the values represent the correlation with paired heads. The averaged correlation is sufficiently high at approximately 0.80, supporting the close link between the RTW attention and self-attention patterns.

\begin{table}[tb]
  \begin{center}
    \resizebox{86mm}{!}{
    \begin{tabular}{l|c|c|c|c|c|c|c|c|c|c|c}
      \hline
      \hline
       \makebox[8mm]{Input} & \makebox[4mm]{$\mathbf{u}_1$} & \makebox[4mm]{$\mathbf{u}_2$} & \makebox[4mm]{$\mathbf{u}_3$} & \makebox[4mm]{$\mathbf{u}_4$} & \makebox[4mm]{$\mathbf{u}_5$} & \makebox[4mm]{$\mathbf{u}_6$} & \makebox[4mm]{$\mathbf{u}_7$} & \makebox[4mm]{$\mathbf{u}_8$} & \makebox[4mm]{$\mathbf{u}_9$} & \makebox[4mm]{$\mathbf{u}_{10}$} &  \makebox[4mm]{Ave.}\\        
      \hline
        1-View 
        & .85 & .85 & .84
        & .84 & .83 & .80 & .80
        & .77 & .74 & .68 & .80 \\    
      \hline
      \hline
    \end{tabular}
    }
  \end{center}
  \caption{Average correlations between RTW attention and self-attention patterns of each pair of canonical angle and head across 174 classes.} 
  \label{tab:cos_sim}
\end{table}

\subsection{Comparison of Video Classification Performance}
\label{sec:comp_acc}
In the previous section, we have shown the close connection between RTW attention and self-attention patterns both qualitatively and quantitatively. 
However, we shall demonstrate that their effectiveness becomes different when applying them to a task of video classification.
For fairness, the sizes of the input sequences $\{\mathbf{I}_i\}$ and $\{{\mathbf{z}}_j\}$ are both fixed at 16. 
The Transformer analyzed 10 views (video clip of 16 frames) extracted evenly by dividing the entire sequence into a specified number of segments, and the output was obtained by integrating the predicted logits from each view.
For RTW, we generated a set of 10 TE features $\{\mathbf{f}_i\}_{i=1}^{10} \in \mathbf{R}^{16}$ by repeatedly performing random sampling. The number of frames $N$ was also 16. The dimension $m_x$ and $m_y$ of hypo subspace were set to 10. These settings allow for sufficient adaptability to the varying lengths of videos in SSv2.

Each model is equipped with a mechanism that allows it to combine the features held by each attention pattern to achieve further effects. Multi-head Self-Attention integrates the discriminative feature vectors $\{\mathbf{r}_j\}$, which reflect the self-attention weights of each head, using a learnable transformation matrix $\mathbf{W}_O$. As mentioned in \cref{sec:rtw}, RTW defines the similarity for the $c$-th class as the average of the top $k$ similarities between the input hypo subspace and multiple reference hypo subspaces $\{{\mathcal{S}}_c^i\}$. The number of reference subspaces for each class is determined according to the amount of class data $r_c$ used to construct them. In this experiment, to observe the combinatorial effects of attention patterns, we fixed $k$ at 10 and varied $r_c$ to 1, 2, 3, and 5.
We also evaluated the performance at one-fifth and one-tenth of the SSv2 dataset to confirm the robustness against the amount of the training data.

\cref{tab:result_acc} compares the two methods regarding recognition rate. From this result, we can see that the RTW outperforms the Transformer without depending on the number of learning videos $r_c$. We understand that this superiority of RTW is due to the property we discussed in the introduction: the RTW attention sees the relationship among all the sequential patterns, while self-attention checks only the relationship within a local view of the sequential patterns. 

Moreover, the performance of RTW using the 10\% subset of the complete dataset is almost identical to that of the Transformer using the full dataset. This finding highlights the clear advantage of RTW over the Transformer when working with a small sample size.

\begin{table}[t]
  \begin{center}
    \resizebox{86mm}{!}{
    \begin{tabular}{l|c|c|c}
      \hline
      \hline
        \makebox[36mm]{Architecture} & \multicolumn{3}{c}{Classification accuracy at three data fractions} \\
            \cline{2-4}      
        & \makebox[18mm]{Full Data}  & \makebox[18mm]{1/5 Data} & \makebox[13mm]{1/10 Data} \\  
      \hline
        Transformer \cite{vivit} & 46.84 & 47.53 & 45.63    \\
      \hline
        RTW $r_c=1$ & 51.06 & 48.29 & 46.74   \\   
        RTW $r_c=2$ & 51.18 & 48.90 & 47.30   \\   
        RTW $r_c=3$ & \underline{51.34} & 49.00 & 47.65   \\   
        RTW $r_c=5$ & 51.11 &  \underline{49.10} & \underline{47.82}  \\ 
      \hline        
      \hline
    \end{tabular}
    }
  \end{center}
  \caption{Classification accuracy (\%) of different methods on Something-Something V2 dataset (174 classes). The parameter $r_c$ is the number of the sets of TE features used to generate one reference hypo subspace.} 
  \label{tab:result_acc}
\end{table}

\section{Conclusion}
\label{sec:conclusion}
In this paper, we revealed that Randomized Time Warping (RTW) can be regarded as a type of self-attention module which is a core component of Transformers. Our research indicated a significant correlation between the attention patterns of RTW and self-attention, despite their different approaches. Furthermore, we discussed that RTW can be applied across all input sequential patterns, while a self-attention module focuses on a clipped local view of the pattern. In performance comparisons using the Something-Something V2 dataset for motion recognition, we confirmed that RTW achieves superior performance over Transformers, demonstrating its advantages. Our findings suggest that RTW could be a viable alternative to the self-attention module.

\bibliographystyle{IEEEbib}
\bibliography{refs}

\end{document}